\pdfoutput=1

\documentclass[11pt]{article}
\usepackage[]{ACL2023}
\usepackage{times}
\usepackage{latexsym}
\usepackage[T1]{fontenc}
\usepackage[utf8]{inputenc}
\usepackage{microtype}
\usepackage{inconsolata}
\usepackage{graphicx}
\usepackage{cuted}
\usepackage{float}
\usepackage{algorithm2e}
\usepackage{amsmath}

\title{Closer Look at Efficient Inference Methods: A Survey of Speculative Decoding}

\author{Hyun Ryu\Thanks{ Equal contribution.}
 \\
  KAIST \\
  \texttt{hyunr01@kaist.ac.kr} \\\And
  Eric Kim\footnotemark[1]\\
  Georgia Institute of Technology \\
  \texttt{ekim625@gatech.edu} \\}
  
\begin{document}
\maketitle

\begin{abstract}

Inference in Large Language Models (LLMs), such as those used in GPT-3 and LaMDA, has relied heavily on autoregressive decoding, which has yielded effective results. However, with LLMs growing in size and complexity, so has the need for improving inference efficiency. The primary bottleneck in this process arises from the computational overhead caused by generating each token sequentially in autoregressive decoding. Speculative decoding has emerged as a promising solution to overcome this bottleneck. Unlike regular autoregressive decoding, which generates one token at a time using a single model, speculative decoding introduces a two-stage process: drafting and verification. In the drafting phase, a preliminary sequence of tokens is generated rapidly in parallel using a smaller, more efficient model. The verification phase then refines this draft, ensuring that the final output aligns with the larger, more sophisticated models. The drafting and verification processes run in parallel to improve efficiency. This paper provides a comprehensive survey of speculative decoding, starting by introducing its fundamental principles and how it was developed to address the efficiency bottlenecks in LLM inference. We will explore various speculative decoding implementations and categorize them into two groups: draft-centric and model-centric. Draft-centric methods focus on finding and verifying the most optimal tokens from a given draft, while model-centric methods focus on improving the quality of draft generation. Finally, we will address the challenges of applying speculative decoding to real-world scenarios. Challenges include throughput, long context generation, model parallelism, hardware limitation, and generalizability. Through this survey, we aim to provide a comprehensive understanding of speculative decoding, its current methods and applications, challenges in real-world deployment, and the potential avenues for future research. 

\end{abstract}

\section{Introduction}
Large Language Models (LLMs) such as GPT-3 \citep{DBLP:journals/corr/abs-2005-14165} and LaMDA \citep{thoppilan2022lamdalanguagemodelsdialog} have shown success in generating coherent and contextually relevant text. However, the impressive capabilities of these models come with significant computational costs, particularly during the inference stage, where text is generated one token at a time. This process, known as autoregressive decoding, requires a sequential evaluation of each token, which becomes increasingly resource-intensive as model sizes grow \citep{kaddour2023challengesapplicationslargelanguage}.

As LLMs continue to scale, the limitations of sequential autoregressive decoding become more pronounced. The larger the LLMs become, the more model parameters there are. The growing number of parameters demands more computational power as the memory access to the parameters becomes the main issue of latency rather than the arithmetic operations \citep{shazeer2019fasttransformerdecodingwritehead}. This memory bottleneck stemming from increased memory access to LLM parameters has driven researchers to seek more efficient decoding methods that can reduce the time and resource required for inference without compromising the quality of the output \citep{shi2024thoroughexaminationdecodingmethods}.

One promising approach that emerged was speculative decoding, which leverages concepts from speculative execution in computer architecture. Speculative execution allows processors to predict the path a program will take and execute instructions along that path before the actual outcome is known. If the prediction is correct, the results of this speculative execution are used, thereby saving time that would have been spent waiting for the decision point \citep{Hennessy_Patterson_2012}. Similarly, speculative decoding involves generating a draft sequence of tokens using a smaller, faster model which operates in parallel with a larger, more accurate model that verifies the draft tokens. The combination of a smaller and larger model effectively balances LLM inference speed and accuracy \citep{leviathan2023fastinferencetransformersspeculative}.

\begin{minipage}{\linewidth}
       \makeatletter
       \@twocolumnfalse
       \makeatother
\begin{algorithm}[H]
\caption{Parallel Speculative Decoding Using a Smaller and Larger Model}
\KwIn{%
    $M_{\text{large}}$: target model (larger model), \\
    $M_{\text{small}}$: auxiliary model (smaller, faster model), \\
    $\alpha$: acceptance threshold, \\
    $N$: maximum sequence length
}
\KwOut{Generated sequence $S = [s_1, s_2, \ldots, s_L]$}
\BlankLine
\textbf{Initialize:} $S \leftarrow [\text{start\_token}]$ \tcp*[r]{Begin with start token}
\For{$i = 1$ \textbf{to} $N$}{
    \tcp{Step 1: Generate candidate tokens with $M_{\text{small}}$}
    Generate a set of candidate tokens $T_{\text{small}} = \{t_1, t_2, \ldots, t_k\}$ from $M_{\text{small}}(S)$ \;
    
    \tcp{Step 2: Evaluate candidates in parallel with $M_{\text{large}}$}
    Compute acceptance probabilities for all $t \in T_{\text{small}}$ in parallel:
    \[
    p_{\text{accept}}(t) \leftarrow \frac{M_{\text{large}}(t | S)}{M_{\text{small}}(t | S)}, \quad \forall t \in T_{\text{small}}
    \]
    
    \tcp{Step 3: Select the first acceptable token}
    \If{any $t$ in $T_{\text{small}}$ has $p_{\text{accept}}(t) \geq \alpha$}{
        Choose the first $t$ with $p_{\text{accept}}(t) \geq \alpha$ and append $t$ to $S$ \;
    }
    \Else{
        Sample token $t_{\text{large}} \sim M_{\text{large}}(S)$ \tcp*[r]{Fallback to target model}
        Append $t_{\text{large}}$ to $S$
    }
}
\end{algorithm}
\end{minipage}

The introduction of speculative decoding addresses the inefficiencies of traditional autoregressive methods by enabling parallel token generation. This approach reduces the overall time required for inference, making it particularly useful for applications that demand real-time or near-real-time processing. Unlike traditional autoregressive decoding that requires K iterations of the model to generate K tokens, speculative decoding reduces the need for constant full-model passes since tokens can be computed in parallel \citep{leviathan2023fastinferencetransformersspeculative}. In addition, this allows the model to access the LLM parameters significantly less, alleviating the memory bottleneck present in previous inference methods.

Among the various speculative decoding methods that have sprung up, a couple have been instrumental in its success and continued advancement. One such method is Medusa where it uses extra decoding heads to process series of subsequent tokens in parallel \citep{cai2024medusasimplellminference}. Another recent method is EAGLE-2 which is able to improve speculative sampling through dynamic draft trees that can look at the context of the model \citep{li2024eagle2fasterinferencelanguage}.

While speculative decoding is a promising step forward towards more efficient LLM inference, it is not without its challenges. Currently, one of the primary concerns is the generalizability of this technique across different tasks. While there are different implementations of traditional speculative decoding, their effectiveness can vary depending on the specific task at hand. For instance, it may accelerate text generation, but not offer the same level of improvement in machine translation \citep{xia2024unlockingefficiencylargelanguage}. This variability highlights the importance of ensuring that advances in inference speed do not come at the cost of performance consistency across various NLP tasks. To truly optimize LLMs, it is crucial to develop techniques that provide uniform improvements across diverse applications, ensuring that the benefits of speculative decoding extend beyond isolated use cases and contribute to improving real-world deployments.

\begin{figure*}[h]
\centering{\includegraphics[scale=.9]{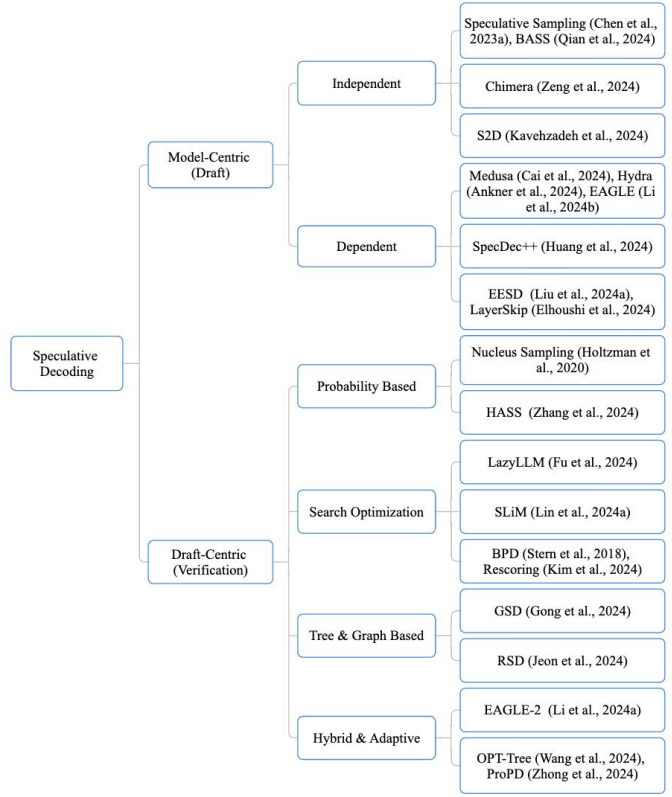}}
\caption{Taxonomy of various speculative decoding methods}
\label{fig}
\end{figure*}

In this paper, we will explore the technical aspects of speculative decoding, including its drafting and verification phases, and examine various implementation strategies. We have separated these strategies into two groups: model-centric implementations and draft-centric implementations. We define model-centric implementations as methods that focus on improving the quality and efficiency of draft generation, typically through changes in the drafter architecture to produce better initial drafts. Draft-centric implementations, on the other hand, focus on selecting a smaller and more refined pool of token candidates, enabling the larger model to verify drafts more efficiently by working through a higher-quality, reduced candidate set. The key difference is that model-centric approaches optimize draft generation itself, while draft-centric approaches refine the candidate pool for verification. We will be talking about existing inference methods such as Medusa among the model-centric implementations and EAGLE-2 among the draft-centric implementations. We will also discuss serving LLMs and the challenges of traditional speculative decoding methods in real-world applications. Finally, we will talk about existing ideas that may help alleviate the issue of real-world deployment and suggest ideas for future research.

\section{What is Drafting?}

The drafting process in speculative decoding is a critical step to accelerate the inference of LLMs by leveraging a smaller, faster model to generate preliminary sequences, or “drafts,” of the text. This process involves running a lightweight model, referred to as the drafting model, in parallel with a larger, more accurate model that is typically used to verify the drafts. The drafting model generates a sequence of K tokens based on a prefix, where K is the number of tokens to generate in a single pass. These drafts may not be perfect but are intended to cover a range of likely outcomes with sufficient accuracy to allow the larger model to quickly verify and fix them if necessary.

Selecting or designing a draft model is crucial as one must consider the trade-offs between speed and accuracy. The Effective Decoding Length (EDL), the length of accepted tokens after drafting \citep{zhao2024lookaheadinferenceaccelerationframework}, captures this trade-off. The goal is to maximize EDL in order to increase inference speed while maintaining accuracy. Using a larger draft model can increase the accuracy of drafts, which would lead to fewer verification steps at the cost of some computational overhead. On the other hand, using a smaller draft model may increase speed and decrease accuracy, which would lead to more runs required by the target model. Another challenge when designing a draft model includes alignment between the draft and target model. More tokens are likely to be accepted by the target model if the draft model aligns well with the target model; in other words, if the target and draft models have similar prediction behaviors. Several studies aim at targeting these challenges by using an independent or dependent drafter model architecture.

\subsection{Model-Centric Implementations}

We categorize speculative decoding methods that focus on improving the quality and efficiency of draft generations as model-centric implementations. These methods are related to the drafter architecture, and sub-categorized into dependent and independent drafter models.

\subsubsection{Independent Draft Models}

Independent drafter models are comparable to the traditional speculative decoding method of using an external, smaller draft model. This draft model can generate drafts quickly due to the smaller number of parameters at the cost of a reduction in accuracy. Using a smaller model of the same series as the target model (e.g. OPT 125M and OPT 7B) is a method known as speculative sampling \citep{xia2024unlockingefficiencylargelanguage}. This approach is effective since models of the same series go through similar pretraining and thus have similar prediction behavior. A similar prediction behavior means more tokens are likely to be accepted during the verification process, consequently reducing the number of forward passes \citep{qian2024bassbatchedattentionoptimizedspeculative}.

Some methods construct a separate draft model through architectural adjustment and model training. For example, Chimera designs its lightweight draft model by creating and training a model with a trigram encoder and a full context encoder \citep{zeng2024chimeralosslessdecodingmethod}. A trigram encoder is used to extract short-range dependencies in the initial layers to reduce inference time. The full context encoder is used for long-range dependencies since a lightweight draft model cannot have multiple transformer blocks.

S2D on the other hand, tackles the challenge of constructing the ideal draft model for a target model \citep{kavehzadeh2024s2dsortedspeculativedecoding}. The approach uses Sorted fine-tuning (SoFT) to create sub-models of a draft model and trains the sub-models on different tasks. It then adaptively selects the ideal sub model that will function as the draft model. These sub models are meant to be versatile for different target models which were created using only one draft model.

To help draft models better align with the target model, some created a framework for training the draft model based on a distillation dataset from the target model \citep{goel2024directalignmentdraftmodel}. Online speculative decoding also addresses alignment between the target and draft model by constantly adjusting the draft model based on the query distribution \citep{liu2024onlinespeculativedecoding}. The main goal is for the draft model to dynamically learn from the target model for better alignment.

\begin{figure*}[h]
\centering{\includegraphics[scale=.52]{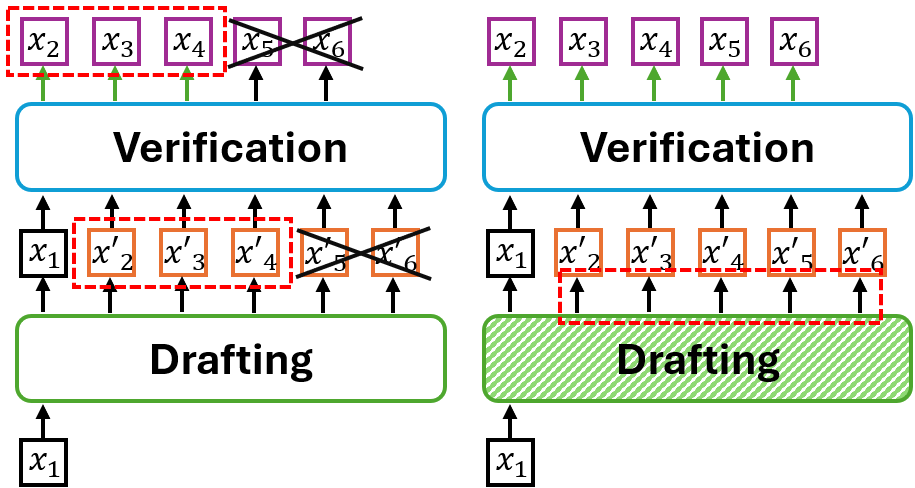}}
\caption{On the left is a general draft-centric implementation that shows the focus on selecting from a smaller pool of drafted tokens compared to standard speculative decoding methods. On the right is a general model-centric implementation that shows that a refined drafting model is used to create better, higher quality initial draft outputs.}
\label{fig}
\end{figure*}

\subsubsection{Dependent Draft Models}

Using an independent drafter model is not always feasible since a smaller model of the same series is not always available and an auxiliary model requires alignment procedures such as knowledge distillation or pre-training. Therefore, using a single model that can accomplish both drafting and verification has gained traction.

The two most common methods in dependent drafting is through adding additional draft heads to the last hidden layer or reducing the number of layer usage. Medusa has shown success by adding multiple heads, which allows the target model to generate multiple tokens without the need of an additional model \citep{cai2024medusasimplellminference}. Hydra has built upon Medusa heads such that each additional head also considers the previously speculated tokens within the continuation \citep{ankner2024hydrasequentiallydependentdraftheads}. This has allowed Hydra to have a higher average acceptance length, meaning the quality of drafts has increased and less iterations are needed. Meanwhile, EAGLE added an embedding layer, and auto-regression head to predict the last-layer representation vector rather than predicting tokens \citep{li2024eaglespeculativesamplingrequires}. Additional prediction heads have also been used to dynamically adjust the candidate length as shown in SpecDec++ \citep{huang2024specdecboostingspeculativedecoding}.

Other methods such as early exit or layer skip reduces the number of layers used for drafting and uses the remaining layers for verification, allowing a single model to accomplish both drafting and verification. Early-exiting Speculative Decoding (EESD) \citep{liu2024speculativedecodingearlyexitingfaster} and LayerSkip \citep{elhoushi2024layerskipenablingearlyexit} both utilize inference using early exit. However, since early exit uses less layers for inference, accuracy is prone to decrease. EESD handles this by using knowledge distillation while LayerSkip utilizes a loss function to help LM heads understand earlier layers. While using a single model may be more efficient for drafting, such methods to compensate for the reduction in accuracy is needed.

\section{What is Verification?}

The verification process in speculative decoding is where the initial drafts generated by the smaller lightweight model undergo rigorous refinement by the larger, more powerful model. This involves taking the multiple candidate sequences produced during the drafting phase and evaluating them against the true model distribution. The verification model, also known as the target model, processes each draft by calculating the log probabilities of each token and determining the overall likelihood of the sequence. If the sequence aligns well with the model’s expectations, it is accepted and goes on to draft another token. If not, the model rejects it and samples another token from an adjusted distribution. This phase is crucial for ensuring that the model is able to balance speed with quality.

However, the verification process introduces several challenges. One significant issue is the computational overhead involved in verifying multiple drafts, especially when many of them might ultimately be discarded. This can offset some of the speed gains achieved in the drafting phase. Additionally, the verification model must be adept at handling diverse sequences produced by the drafting model, which can vary widely in quality. This requires the verification model to be highly flexible and capable of making nuanced adjustments, which increases its complexity.

To address these challenges, several approaches can be employed. One strategy is to implement more sophisticated filtering mechanisms before the verification phase, reducing the number of drafts that need full evaluation by the large model. For instance, initial filtering could be based on a scoring system that discards low-quality drafts early on. Another approach involves using techniques like dynamic programming or beam search to prioritize the most promising sequences, thereby focusing computational resources on the drafts with the highest potential. Moreover, adaptive verification strategies that adjust the depth of verification based on the perceived quality of the draft could further optimize performance. These methods aim to streamline the verification process, ensuring that it remains efficient while maintaining the high quality of text generation expected from large-scale LLMs.

\subsection{Draft-Centric Implementations}

We categorize speculative decoding methods that focus on choosing from a smaller and better pool of token candidates as draft-centric implementations. This allows for the bigger model to verify drafts more efficiently as it is given a smaller pool of better quality candidates to go through. These methods are sub-categorized into probability-based, search optimization, tree and graph-based, and hybrid and adaptive approaches.

\subsubsection{Probability-Based Approaches}

Probability-based approaches involve manipulating the probability distribution of token candidates to refine the draft pool, focusing on generating sequences with high likelihoods or diverse options based on probabilistic criteria. This is crucial for refining the draft pool, ensuring that the sequences produced are not only probable but also varied enough to cover different possible continuations.

Traditional maximization-based methods like beam search, although effective in many contexts, often lead to text degeneration - producing output that is repetitive, bland, or incoherent. This issue arises because such methods tend to focus too heavily on high-probability tokens, which can cause the model to get stuck in loops or generate predictable text.

To counteract these issues, nucleus sampling has been proposed as an effective alternative \citep{holtzman2020curiouscaseneuraltext}. This method improves upon previous strategies by truncating the probability distribution, excluding the unreliable tail where less probable (and often less meaningful) tokens reside. Instead, nucleus sampling dynamically selects from the nucleus of tokens that represent the bulk of the probability mass, ensuring that the generated text remains coherent and diverse. This approach is able to generate high-quality long-form text and retain the diversity seen in human-written content by balancing the trade-offs between likelihood and diversity.

Other methods opted to optimize the existing speculative sampling instead. Harmonized Speculative Sampling (HASS) presents a refined strategy that aligns the model’s training and decoding processes to enhance efficiency \citep{zhang2024learningharmonizedrepresentationsspeculative}. HASS optimizes speculative sampling by adjusting the probability distributions during the inference stage, which increases the acceptance rate of generated drafts. By harmonizing the probabilities used during training with those applied during sampling, the model produces drafts that are more likely to be accepted without needing extensive recalculations. This alignment reduces computational overhead while maintaining high-quality outputs, making speculative decoding more practical for real-time applications.

\subsubsection{Search Optimization Techniques}

Search optimization techniques focus on systematically narrowing down the search space for token sequences by pruning less promising candidates, ensuring that the most likely or highest-quality sequences are prioritized for verification.

One such technique is LazyLLM \citep{fu2024lazyllmdynamictokenpruning}, which introduces a novel method that optimizes the inference process by dynamically pruning tokens deemed less critical for the generation of subsequent tokens. Unlike static pruning methods that permanently remove tokens from consideration, LazyLLM selectively computes the key-value (KV) pairs for tokens that are important for the next token prediction, deferring the computation of less critical tokens to later steps when they might become relevant. This dynamic approach allows the model to efficiently handle long prompts by progressively reducing the number of tokens involved in each computation, thus speeding up both the pre-filling and decoding stages.

SLiM (Speculate Less, Validate More) \citep{lin-etal-2024-slim} complements these efforts by incorporating a hypothesis reduction stage between the speculative drafting and verification phases. It addresses the computational overhead of verifying numerous speculative tokens by introducing a lightweight verification step that uses fast posterior estimation to eliminate unlikely candidates early in the process. This reduction in the number of tokens requiring full verification significantly cuts down on floating-point operations (FLOPs) and leads to substantial computation savings. SLiM is particularly effective in scenarios with constrained computational budgets while improving on its predictions.

Blockwise Parallel Decoding (BPD) introduces a method focused on generating blocks of multiple tokens simultaneously to accelerate decoding in LLMs \citep{stern2018blockwiseparalleldecodingdeep}. A key challenge of BPD lies in ensuring that the predicted blocks are both coherent and fluent. In a draft-centric context, the BPD approach is particularly relevant, as it aims to refine the draft pool to include smaller but higher-quality blocks of token candidates. The introduction of rescoring mechanisms, such as n-gram models and neural autoregressive models, helps to prune these drafts by rescoring the top-k block predictions, thereby enhancing block efficiency \citep{kim2024exploringimprovingdraftsblockwise}. This pruning aligns with the draft-centric philosophy, which emphasizes selecting a better and smaller candidate pool for verification. By improving BPD’s inference speed with the pruning algorithm, it is able to optimize the speculative decoding process, ultimately improving both the efficiency and accuracy of the verification step.

\subsubsection{Tree and Graph-Based Methods}

Tree and graph-based methods utilize tree or graph structures to explore and optimize the selection of token candidates. They focus on balancing exploration and exploitation to refine the draft pool systematically, thereby optimizing the draft pool and enhancing the overall efficiency of LLMs.

Graph-Structured Speculative Decoding (GSD) \citep{gong2024graphstructuredspeculativedecoding} introduces a novel approach that extends traditional tree-based methods by incorporating a directed acyclic graph (DAG) to manage the drafted token sequences. Unlike tree-based structures, where each hypothesis expands independently, GSD recognizes that many hypotheses share common token sequences. By merging these common sequences into a graph structure and pruning unlikely tokens, GSD reduces redundancy and computational load, allowing the model to process a broader range of hypotheses more efficiently. This method significantly enhances the acceptance rate of drafted tokens while minimizing the computational overhead typically associated with regular tree-based speculative decoding.

Recursive Speculative Decoding (RSD) \citep{jeon2024recursivespeculativedecodingaccelerating} further expands on tree-based methods by introducing a recursive approach that maximizes the diversity of draft tokens while minimizing computational overhead. RSD utilizes a tree structure to organize draft tokens, employing sampling without replacement techniques such as the Gumbel-Top-k trick and Stochastic Beam Search. These techniques allow RSD to build a more diverse and effective draft-token tree, which is particularly beneficial for scenarios where computational resources are limited. The method also incorporates early truncation of unlikely sequences, reducing the overall computational cost and improving efficiency compared to previous tree-based speculative decoding approaches.

\begin{figure*}[h]
\centering{\includegraphics[scale=.7]{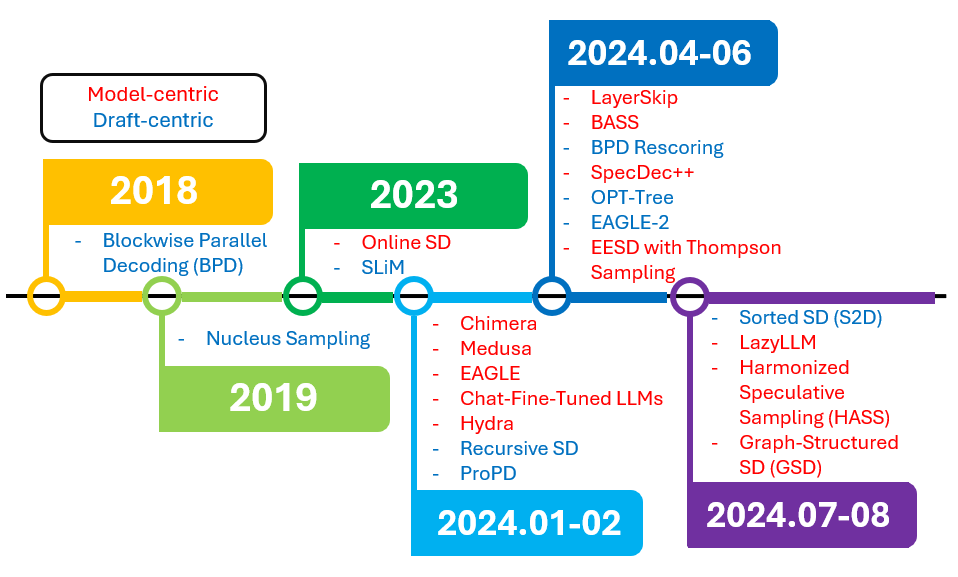}}
\caption{Timeline of the various speculative decoding methods discussed in this paper}
\label{fig}
\end{figure*}

\subsubsection{Hybrid and Adaptive Approaches}

Hybrid and adaptive approaches include methods that combine multiple strategies or adapt the drafting process based on specific criteria like context or input conditions to generate more refined and task-specific drafts, making the verification process more efficient.

EAGLE-2 is an advanced tree-based speculative decoding method that dynamically adjusts the draft tree structure based on context \citep{li2024eagle2fasterinferencelanguage}. It departs from the static tree structures used in traditional speculative decoding by introducing a context-aware, dynamic draft tree. This method uses confidence scores from the draft model to approximate the acceptance rates of tokens, allowing for a more efficient and targeted expansion of the draft tree. The draft tree is selectively grown by focusing on the most promising branches, reducing the number of low-quality token candidates early in the process. This draft-centric approach significantly enhances the efficiency of the verification stage, reducing the need for recomputation and improving inference speeds. By leveraging tree structures that adapt to token acceptance rates, EAGLE-2 is able to optimize the balance between exploration and exploitation, ensuring better verification of token sequences while maintaining high computational efficiency.

OPT-Tree \citep{wang2024opttreespeculativedecodingadaptive} exemplifies an advanced hybrid approach. It dynamically constructs and adjusts a draft tree structure during each decoding step. This method is designed to maximize the mathematical expectation of the acceptance length, which directly correlates with the efficiency of the decoding process. The adaptive nature of the tree structure allows it to respond to different input sequences and computational resources, making it highly scalable and effective across various scenarios. OPT-Tree’s ability to generate multiple tokens in a single decoding step while maintaining a high acceptance rate significantly improves the speed and efficiency of LLMs.

Similarly, ProPD \citep{zhong2024propddynamictokentree} incorporates both early pruning and dynamic tree generation strategies to enhance speculative decoding. ProPD’s early pruning algorithm efficiently reduces the number of token candidates that need to be verified, focusing computational resources on the most promising sequences. This is followed by a dynamic tree generation algorithm that adjusts the token tree structure in real-time based on current decoding conditions, such as batch size and sequence length. The combination of these strategies allows ProPD to achieve substantial speedups in parallel decoding tasks, demonstrating superior performance across different models and tasks.

\section{Applications \& Future Challenges}

When it comes to serving LLMs in real-world applications, traditional speculative decoding methods fall short in scenarios beyond simple speed-up optimization. These methods often focus exclusively on reducing token generation latency, but real-world deployment requires balancing multiple factors, such as system load, computational overhead, and the ability to handle high-throughput environments effectively. Speculative decoding’s additional computational burden can increase latency under high request rates or low token acceptance accuracy \citep{liu2024optimizingspeculativedecodingserving}. This inefficiency is compounded when speculative decoding is applied without considering the overall system’s capacity, leading to bottlenecks rather than improvements in latency.

In the following sections, we will address several key challenges in deploying LLMs effectively. These include throughput, where the focus is on ensuring that systems can handle large volumes of requests efficiently; long context generation, which addresses the difficulties in generating coherent responses over extended interactions; hardware limitations, which focuses on memory and computational restrictions of models; model parallelism, which examines how speculative decoding methods can better distribute workload across multiple hardware units; and generalizability, ensuring that speculative decoding methods provide consistent speed-up across various tasks and use cases. These topics will explore how speculative decoding can be adapted to meet the demands of real-world deployment, moving beyond mere speed-up to more holistic performance improvements.

\subsection{Throughput}

Throughput, in the context of serving LLMs, refers to the system’s ability to handle multiple requests or generate tokens quickly, especially in scenarios where high volumes of data need to be processed in real-time. One of the major challenges in achieving high throughput with LLMs is balancing the need for low latency with maintaining model accuracy and performance across multiple tasks. Traditional speculative decoding techniques focus primarily on reducing latency by speeding up token generation, but they often neglect throughput when serving LLMs in real-world, high-demand environments.

MagicDec addresses one of the key limitations of speculative decoding by tackling the latency-throughput trade-off in long-context applications, such as interactive chatbots and document processing \citep{chen2024magicdecbreakinglatencythroughputtradeoff}. Traditional speculative decoding techniques have struggled with throughput in larger batch sizes, as these methods were typically optimized for small batches or single sequences. MagicDec adaptively manages the KV (key-value) cache associated with token generation where the size of the KV cache grows and becomes a bottleneck as sequence length and batch size increase. MagicDec solves this by using sparse KV caches, which allow more sequences to be handled simultaneously without overwhelming computational resources. This approach enables MagicDec to deliver speedups even in high-throughput environments, where traditional speculative decoding techniques would falter.

BASS (Batched Attention-optimized Speculative Sampling) goes beyond traditional speculative decoding by optimizing both latency and throughput in a batched setting, which is crucial for real-world applications that require handling multiple responses simultaneously \citep{qian2024bassbatchedattentionoptimizedspeculative}. Unlike single-sequence speculative decoding methods, BASS is designed to perform well in a batched environment, where multiple token sequences need to be generated in parallel. It achieves this through an optimized attention mechanism that ensures efficient GPU utilization and minimizes the computational overhead typically associated with handling large batches of data. This method not only accelerates token generation per sequence but also optimizes the system’s ability to handle high-demand environments, making it particularly suited for applications like interactive AI systems, real-time content generation, and multi-tasking LLM deployment.

Several key areas stand out when considering the future challenges for throughput in speculative decoding. First, scalability remains a significant challenge, particularly as LLMs grow in size and complexity. While techniques like MagicDec and BASS have demonstrated success in optimizing throughput currently, future models with more parameters and longer context windows will require increasingly efficient decoding strategies. Ensuring that throughput scales effectively with model size, without leading to diminishing returns in performance, will be an important area of research to consider.

Another challenge lies in balancing throughput with energy efficiency. High-throughput speculative decoding methods often require increased computational resources, leading to higher energy consumption. As LLMs become more widely deployed, particularly in resource-constrained environments like edge computing or mobile devices, the trade-off between maximizing throughput and minimizing energy usage will become more pronounced. Research into energy-efficient hardware accelerators, as well as optimized algorithms, will be crucial to address this challenge.

\subsection{Long Context Generation}

Long-context generation is the ability of LLMs to handle and generate text that extends over lengthy inputs, such as in applications like document analysis or interactive chatbots. The main challenge in long-context generation arises from the increased computational and memory overhead associated with managing and processing long sequences. As the context window expands, models must retain and utilize information over many tokens, which often leads to bottlenecks in performance, particularly in the management of the KV (key-value) cache. This cache grows substantially with longer inputs, making it more difficult to maintain low latency and high accuracy.

Recent techniques like MagicDec address these issues by optimizing the speculative decoding process for long-context scenarios \citep{chen2024magicdecbreakinglatencythroughputtradeoff}. It adapts its KV cache management, using sparse caching strategies to alleviate the computational load. This not only helps handle longer sequences but also optimizes throughput, as discussed in the throughput section. By dynamically managing cache usage, MagicDec offers significant improvements for long-context applications without sacrificing generation quality.

In addition, approaches like TriForce tackle long-context generation by leveraging attention sparsity within the model \citep{sun2024triforcelosslessaccelerationlong}. TriForce uses hierarchical speculative decoding and cache eviction strategies, such as the heavy-hitters approach, to selectively retain critical tokens and discard less important ones. This allows models to maintain performance over extremely long contexts (e.g., 120K tokens) while reducing memory overhead. Such strategies highlight the increasing focus on managing computational resources effectively to ensure that LLMs can handle extensive inputs efficiently.

Looking ahead, future challenges in long-context generation will involve not only optimizing memory usage but also improving the generalization of these techniques across different tasks. As models continue to scale and applications demand more extensive context windows, ensuring consistent performance while minimizing latency and memory consumption will be critical areas of research. Additionally, adaptive approaches that can dynamically adjust to varying context lengths and workloads will become increasingly important as real-world use cases evolve.

\subsection{Model Parallelism}

To maximize model parallelism, the GPU needs to be used to its full extent. When serving LLMs in real world scenarios, multiple users will be interacting simultaneously and their requests should be grouped together in batches to maximize GPU utilization. This technique, known as “batching” is generally not studied in speculative decoding models, as most models examine single sequence generations. Batching increases model parallelism, since multiple sequences can be processed. However, having a variable number of tokens in a batch can be problematic, since the sequences take a varying amount of time to complete. This means the GPU wastes resources waiting for the longest sequence in the batch to complete generation \citep{nie2024aladdinjointplacementscaling}.

BASS tackles this problem of variable token length by padding the KV cache to match the length of the longest sequence within a batch. However, this can lead to extra memory overhead and unnecessary computations being performed. Therefore, BASS also proposes a split method where the K, V, P tensors are broken into smaller sequences to handle different sequence lengths \citep{qian2024bassbatchedattentionoptimizedspeculative}. BASS is able to achieve a max GPU utilization of 15.8\%, which is 10 times greater compared to vanilla speculative decoding method.

Efficient Multi-sample Speculative Decoding (EMS-SD) acknowledges the memory overhead and wasted computation of padding tokens and proposes using a second “unpad” KV cache that holds the location of the KV cache of different batches \citep{ni2024emssdefficientmultisamplespeculative}. During the verification phase, the target model can use the unpad KV cache to find the start location of each sequence, which means the target model can accept sequences of varying lengths. With this method, there is a more gradual decline in speed up as batch size increases compared to the vanilla method. The speedup is 2.85x and 0.88x for batch sizes of 1 and 16 respectively for BASS, while the speedup is 2.50x and 0.48x for the vanilla method.

While the main challenge for batch speculative decoding is the variable token length of batches, there are other challenges such as token rejection and GPU load balancing that might limit parallelism gains. If a significant number of tokens are rejected after the drafting phase, this can lead to inefficiencies in a batch, as the errors must be corrected while still processing the remainder of the batch. Moreover, not all sequences require the same amount of GPU usage, since more complex sequences require more compute power. Additional research on such challenges of batching will be beneficial to maximize model parallelism.

\subsection{Hardware Limitation}

In real world applications, people do not have the computational resources to perform extensive training or inference of the speculative decoding models. This is apparent when using an independent drafter model, as the drafter model needs to be trained to align with the target model. Meanwhile, some speculative decoding methods that do not use a drafter model are bound by memory due to the complexity of additional modules or large token trees.

Parallel Prompt Decoding (PPD) tackles the memory overhead problem of token trees by using a hardware-aware dynamic sparse tree technique. This method adjusts the prompt structure during runtime based on the forward pass latency of different hardware types. By dynamically adjusting the tree to have different depth and structure based on a given hardware, memory resource is allocated optimally. As a result, PPD has a memory overhead that is 0.004\% compared to that of Medusa and 0.007\% compared to Eagle \citep{chen2024hardwareawareparallelpromptdecoding}. PPD also trains efficiently by training the embeddings of prompt tokens instead of a new model. This training method requires only 0.0002\% additional training parameters, compared to 8.07\% from Medusa.

In some cases where memory is limited, quantization may be needed, which can lead to a significant slow down of up to 7 times \citep{lin2024qservew4a8kv4quantizationcodesign}. Reducing quantization overhead is especially crucial when using HBM VRAM due to its high cost. Skippy Simultaneous Speculative Decoding (S3D) aims at improving speculative decoding on low memory GPUs through midlayer skipping and multi-token prediction. By skipping layers, S3D utilizes an independent drafting method and saves memory by eliminating the need for an extra model and additional training for the draft model. S3D achieved a slightly lower VRAM usage of 8.06 GiB compared to 9.63 GiB from EAGLE without sacrificing model speed.

Since computational power and resources vary significantly across different hardware, it is important to develop models that can cater to different needs. Most studies focus on using the most powerful GPU for training, but this is not a feasible approach. Models like PPD where the algorithm adjusts dynamically to the hardware is a step in the right direction. Recently, parameter offloading has gained popularity, where the model parameters are stored in a CPU RAM and loaded into the GPU during inference \citep{svirschevski2024specexecmassivelyparallelspeculative}. However, this leads to the inherent problem of quantization overhead as mentioned in S3D and more research on tackling quantization overhead for weaker devices will be needed.

\subsection{Generalizability}

While most speculative decoding models are good at a specific task, no model excels at all tasks, meaning it is not generalizable. SpecBench is a commonly used benchmark to compare speedup of different speculative decoding models across different tasks. These tasks include multi-turn conversation, translation, summarization, question answering, mathematical reasoning, and retrieval-augmented generation \citep{xia2024unlockingefficiencylargelanguage}. Spec-Bench has been used to compare speedups of a handful of methods like Lookahead, Medusa, and EAGLE. Due to EAGLE’s success in nearly every task, EAGLE has commonly been used as the de facto model for comparison and there have been many models since that have greater speedups on individual tasks.

However, in real world scenarios, it may not be feasible to utilize individual models for each specific task required, as this would introduce complexities and costs of managing multiple models. Managing multiple models can lead to latency from switching between models and maintaining consistency across models. Therefore, it is important to continue research towards a versatile model that could find the greatest speed up across most tasks.

\section{Conclusion}

In this paper, we examine different speculative decoding approaches and categorize them into draft-centric and model-centric implementation groups. Model-centric approaches are concerned with drafting tokens of quality while draft-centric approaches focus on efficiently selecting and verifying from the draft tokens. We then analyze serving speculative decoding in the real world and the challenges it faces: throughput, long context generation, model parallelism, hardware limitation, and generalizability. We believe speculative decoding is a promising approach for LLM inference optimization, but addressing the limitations will be crucial in order to apply it in the real world.

\bibliography{custom}
\bibliographystyle{acl_natbib}

\end{document}